\documentclass[letterpaper]{article} % DO NOT CHANGE THIS
\let\svthefootnote\thefootnote
\newcommand\blankfootnote[1]{%
  \let\thefootnote\relax\footnotetext{#1}%
  \let\thefootnote\svthefootnote%
}
\usepackage{aaai23}  % DO NOT CHANGE THIS
\usepackage{times}  % DO NOT CHANGE THIS
\usepackage{helvet}  % DO NOT CHANGE THIS
\usepackage{courier}  % DO NOT CHANGE THIS
\usepackage[hyphens]{url}  % DO NOT CHANGE THIS
\usepackage{graphicx} % DO NOT CHANGE THIS
\def\UrlFont{\rm}  % DO NOT CHANGE THIS
\usepackage{natbib}  % DO NOT CHANGE THIS AND DO NOT ADD ANY OPTIONS TO IT
\usepackage{caption} % DO NOT CHANGE THIS AND DO NOT ADD ANY OPTIONS TO IT
\usepackage{algorithm}
\usepackage{algorithmic}

\usepackage{newfloat}
\usepackage{listings}
\DeclareCaptionStyle{ruled}{labelfont=normalfont,labelsep=colon,strut=off} % DO NOT CHANGE THIS
\floatstyle{ruled}
\newfloat{listing}{tb}{lst}{}
\floatname{listing}{Listing}
\title{Progressive Multi-view Human Mesh Recovery with Self-Supervision}
\author {
    Xuan Gong\textsuperscript{\rm 1,\rm 2},
    Liangchen Song\textsuperscript{\rm 1,\rm 2},
    Meng Zheng\textsuperscript{\rm 1},
    Benjamin Planche\textsuperscript{\rm 1},\\
    Terrence Chen\textsuperscript{\rm 1},
    Junsong Yuan\textsuperscript{\rm 2},
     David Doermann\textsuperscript{\rm 2},
    Ziyan Wu\textsuperscript{\rm 1}
}
\affiliations {
    \textsuperscript{\rm 1} United Imaging Intelligence, Cambridge MA, USA 02140\\\{meng.zheng, benjamin.planche, terrence.chen, ziyan.wu\}@uii-ai.com \\ %\{first.last\}@uii-ai.com\\
    \textsuperscript{\rm 2} University at Buffalo, Buffalo NY, USA 14260\\\{xuangong, lsong8, jsyuan, doermann\}@buffalo.edu
    
}

\usepackage{tikz}
\usepackage{comment}
\usepackage{amsmath,amssymb} % define this before the line numbering.
\usepackage{color}
\usepackage{epsfig}
\usepackage{tabularx}
\usepackage{booktabs}
\usepackage{multirow}
\usepackage{bm}
\usepackage{cite}
\usepackage[switch]{lineno}
\usepackage{xspace}
\usepackage{pifont}% http://ctan.org/pkg/pifont
\usepackage{textcomp}

\newcommand{\textapprox}{\raisebox{0.5ex}{\texttildelow}}

\newcommand{\cmark}{\textcolor{red}{\ding{51}}}%
\newcommand{\xmark}{\textcolor{teal}{\ding{55}}}%

\makeatletter
\DeclareRobustCommand\onedot{\futurelet\@let@token\@onedot}
\def\@onedot{\ifx\@let@token.\else.\null\fi\xspace}

\def\eg{\emph{e.g}\onedot} 
\def\ie{\emph{i.e}\onedot}

\def\wrt{w.r.t\onedot} 

\def\etal{\emph{et al}\onedot}
\makeatother

\newcommand{\Loss}{\mathcal{L}}

\usepackage{rotating}

\begin{document}

\maketitle
%\linenumbers

\begin{abstract}
% Existing 3D human mesh estimators suffer poor generalization performance to new datasets, largely due to the limited diversity of image-3D pose pairs in training data. 
% In the era of deep learning, human pose estimation from
% multiple cameras with unknown calibration has received
% little attention to date. We show how to train a neural
% model to perform this task with high precision and minimal latency overhead.
% In diverse tests, we show that HUND \bnote{what is HUND?} achieves very competitive results in datasets like H3.6M and 3DPW, as well as good quality 3d reconstructions for complex imagery collected in-the-wild. gained considerable attention
To date, little attention has been given to multi-view 3D human mesh estimation, despite real-life applicability (\eg, motion capture, sport analysis) and robustness to single-view ambiguities. Existing solutions typically suffer from poor generalization performance to new settings, largely due to the limited diversity of image-mesh pairs in multi-view training data. To address this shortcoming, people have explored the use of synthetic images. But besides the usual impact of visual gap between rendered and target data, synthetic-data-driven multi-view estimators also suffer from overfitting to the camera viewpoint distribution sampled during training which usually differs from real-world distributions. Tackling both challenges, we propose a novel simulation-based training pipeline for multi-view human mesh recovery, which (a) relies on intermediate 2D representations which are more robust to synthetic-to-real domain gap; (b) leverages learnable calibration and triangulation to adapt to more diversified camera setups; and (c) progressively aggregates multi-view information in a canonical 3D space to remove ambiguities in 2D representations.
Through extensive benchmarking, we demonstrate the superiority of the proposed solution especially for unseen in-the-wild scenarios. 
%on par with fully/weakly-supervised methods. 
\end{abstract}

\section{Introduction}
As a key step to several human-centric applications, 3D human mesh estimation from multi-view images has shown superiority beyond monocular image as it eliminates common ambiguities among single-image scenarios. 
Most successes in 3D human mesh recovery are demonstrated by supervised training. However, such models hardly generalize to in-the-wild scenarios due to the lack of sufficient and diverse 3D annotations paired with multi-view images.
Generalizability of human mesh estimation models developed using supervision on large-scale in-studio datasets remains questionable, as these models often perform unsatisfactorily on unseen in-the-wild environments. Though weakly-supervised models can be an option to address this shortcoming, their performance highly relies on availability and diversity of unpaired 2D/3D annotations or temporal/multi-view image pairs, therefore limiting generalization capability. 

\begin{figure*}[ht!]
	\centering
	\includegraphics[width=\linewidth]{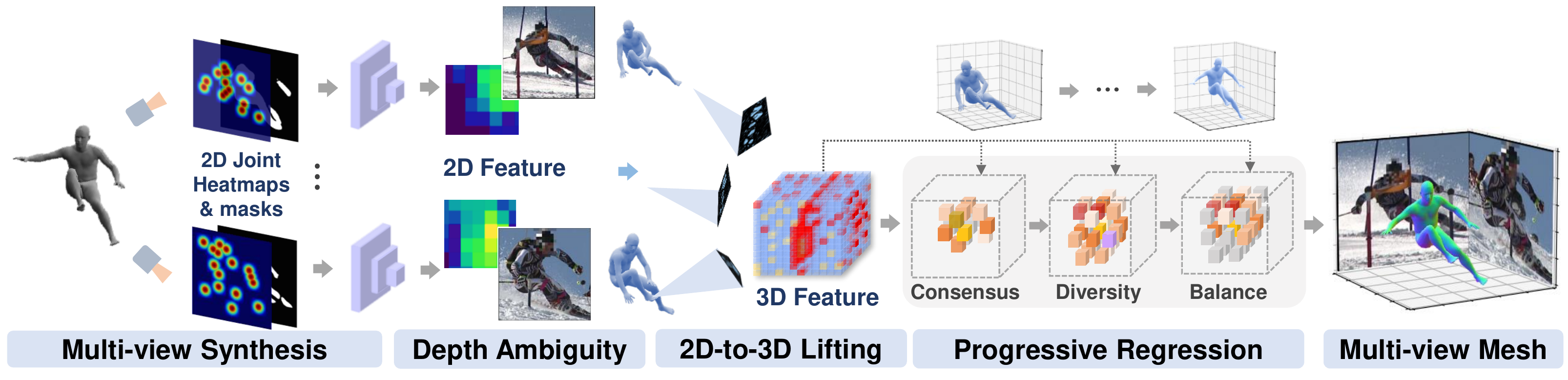}
	\vspace{-2em}
	\caption{(1) Our method is purely synthetic-data-driven regressing human mesh from 2D representations. During training 2D representations are acquired from synthetic mesh rendering. During testing 2D representations are predicted with off-the-shelf detectors. The existing well-trained 2D detectors equip our method with better generalizability and robustness to in-the-wild scenarios. 
	(2) By unifying multi-view knowledge in canonical 3D space, we explicitly explore consensus, diversity and balance to deal with the inherent inconsistency among different views. 
	%We aggregate multi-view 2D representations are in the shared 3D human space and progressively regressed the human mesh. 
	}
	\label{fig:teaser}
	\vspace{-1em}
\end{figure*}
% Despite the recent success of multi-view 3D human mesh reconstruction methods, recovering the accurate 3D human mesh from multi-view images is still challenging due to the lack of suitable training datasets with in-the-wild images paired with accurate and diverse body mesh labels. Learning to regress 3D human body shape and pose (\eg, SMPL parameters) from monocular images typically exploits losses on 2D keypoints, silhouettes, and/or part segmentation when 3D training data is not available. Such supervisions, however, 

\blankfootnote{This work was carried out during the internship of X. Gong and L. Song at United Imaging Intelligence, Cambridge, MA.}

When collected training data are insufficient and infeasible to generalize, simulated training data can be a useful alternative.
Works have been done to synthetically generate and render images of human bodies for dense pose estimation \cite{zhu2020simpose}, depth estimation \cite{varol2017learning}, 3D pose estimation \cite{rogez2016mocap, varol2017learning, kundu2020self, patel2021agora}, 3D human reconstruction \cite{zheng2019deephuman, sengupta2020synthetic, yu2021skeleton2mesh}. Most of the self-supervised approaches only focus on single view tasks. Although works such as \citet{kocabas2019self, wandt2021canonpose} employ multi-view geometry for self-supervised training, for testing they infer on individual images independently, even when multi-view data are available. We believe that there commonly exist multi-view images in real scenarios, but self-supervised multi-view human pose/mesh estimation remains relatively unexplored. 

Some works have taken steps to eliminate the requirements of 3D human mesh annotation.  \citet{pavlakos2017harvesting} explore 3D geometry of the camera setup to lift from multi-view 2D joints to 3D pictorial structure. But the training process highly relies on multi-view imagery and only estimate human pose excluding human shape. \citet{liang2019shape} generate multi-view human RGB images from existing SMPL \cite{loper2015smpl} pose and shape parameters. But its synthetic paired data only helps to improve the performance of model trained with real image-3D annotations. Although these works have deliberately designed human textures, light, background during rendering, the model only trained with its synthetic data can hardly generalize to real scenarios due to the domain gap between real and synthetic images. It is obvious that the rendered images can hardly generalize to the real images in the wild. %\etc, additional efforts have to be done when adapting the model trained with synthetic data to real tasks. 

On the other hand, proxy representations (\eg, joints, silhouettes) are commonly used as intermediate representation in human mesh recovery (HMR) task, and can serve as good transition to bridge RGB image and SMPL parameters: 1) synthetic-to-real domain gap is smaller for proxy representation, thus can be more easily bridged; 2) a vast amount of SMPL parameters can be rendered into proxy representations to formulate paired synthetic data; 3) there exist multiple well-trained models predicting image to these 2D lower-dimensionality representations, which are relatively more robust and generalized compared with 3D mesh predictors.

To this point, we propose to train with multi-view synthetic data for multi-view human mesh estimation. Compared to single-view synthetic training \cite{pavlakos2018learning,sengupta2020synthetic,yu2021skeleton2mesh, gong2022self, zheng2022self}, two challenges specific to multi-view synthetic training arise. First, an additional domain gap between the real testing data and synthetic training data can be easily introduced due to inconsistency between the testing camera viewpoint distribution and the synthetic training viewpoints. Due to the lack of available multi-view SMPL parameters and camera calibration, the sampling of camera setups for multi-view synthesis should generalize to real multi-view settings which can be quite diverse across different datasets. % that can directly be leveraged, 
% While several multi-view datasets provide camera between its defined world coordinate and camera coordinate, the position and global rotation of human body relative to world coordinate remains unknown and can be quite diverse across different datasets. 
The second challenge comes from the inherent inconsistency among multi-view representations in real scenarios. Limited by occlusion and depth ambiguity, 2D representations inferred from testing images are more likely to be biased from 3D ground-truth when compared with images which are used in those fully supervised multi-view methods.

\begin{figure*}[ht!]
	\centering
	\includegraphics[width=\linewidth]{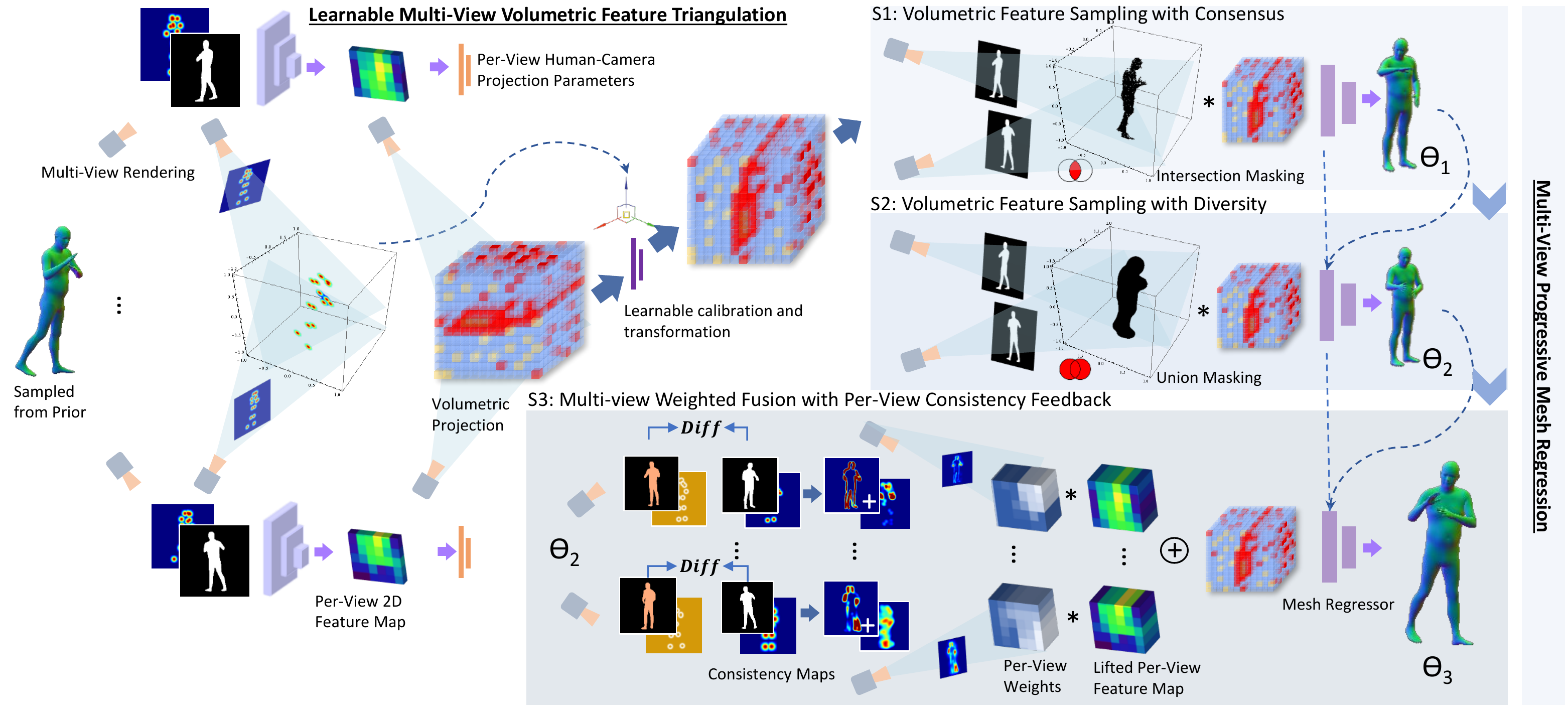}
	\vspace{-2em}
	\caption{Overall illustration of the proposed pipeline. We unproject the 2D occupancy maps to 3D space to obtain intersection and union of 3D occupancy used to mask the volumetric features. The intersection and union masking in 3D space help to sample volumetric features according to multi-view consensus and diversity respectively. This progressive regression design can efficiently
    tolerate possible bias and seek consistency in real multi-view settings.}
	\label{fig:pipeline}
	\vspace{-1em}
\end{figure*}

To address the aforementioned issues, we propose a novel synthetic-data-driven training pipeline (Figure \ref{fig:teaser}) for multi-view human mesh recovery.  1) We synthetically train a regression model from multi-view 2D representations to SMPL parameters. During inference off-the-shelf detection/segmentation models are used to predict these 2D representations from RGB images. 2) The viewpoint setup for synthetic training is consistent with real testing scenarios via learnable volumetric triangulation and calibration. 3) Multi-view 2D representations are aggregated in the shared 3D human space and progressively regressed to deal with the possible bias existing in 2D representation. 
As illustrated in Figure \ref{fig:pipeline}, we aim to let the regressor first focus on the consensual key area and then learn from the possible highlighted area for diversity. Once we get mesh prediction from these two regression iterations, we are able to better balance the multi-view volumetric features via reprojection consistency where views less consistent with the predicted mesh will be given less weight in the final mesh refinement iteration.  %Our synthetically trained model is agnostic to any view number larger than two and generalizes on different benchmark datasets for testing. 
Empirical evaluations show that our method achieves very competitive results on H3.6M \cite{ionescu2013human3}, TotalCapture \cite{Trumble:BMVC:2017}  and challenging SkiPose \cite{sporri2016reasearch,rhodin2018learning} dataset compared with other fully/weakly supervised multi-view human mesh recovery and human pose estimation methods. 

Our key contributions can be summarized as:
1) We propose a multi-view synthetic-data-driven training pipeline for multi-view human mesh recovery, mapping multi-view 2D representations to shared 3D human space to bridge the real-synthetic gap. %To the best of our knowledge, we are the first work doing self-supervised multi-view human mesh estimation. 
2) We progressively regress multi-view representations by first exploring the consensus and diversity among views in 3D space and then reaching evidential balance among views. This design can efficiently tolerate the bias commonly existing in single 2D representation thus generalizable to in-the-wild scenarios. 3) We conduct extensive experiments on standard benchmark datasets and demonstrate comparable numbers  with fully/weakly supervised methods on conventional evaluation metrics.
%  \begin{itemize}
%     \item We propose a multi-view synthetic-data-driven training pipeline for multi-view human mesh recovery, mapping multi-view 2D representations to shared 3D human space to bridge the real-synthetic gap. To the best of our knowledge, we are the first work of self-supervised multi-view human mesh estimation.
    
%     \item We progressively regress multi-view representations by first exploring the consensus and diversity among views in 3D space and then reaching evidential balance among views. This design can efficiently tolerate the bias commonly existing in single 2D representation thus generalizable to in-the-wild scenarios.    
    
%     \item We conduct extensive experiments on standard benchmark datasets and demonstrate comparable numbers  with fully/weakly supervised methods on conventional evaluation metrics.
% \end{itemize}

% \vspace{-1.em}
\section{Related Works}
% \vspace{-.5em}

\noindent \textbf{Monocular 3D Human Pose Estimation:}
3D human pose estimation (HPE) \cite{agarwal2005recovering, song2021human} problem can be categorized into 3D body keypoint/skeleton prediction and 3D human mesh recovery, based on representing the human body with kinematic or volumetric models. 3D keypoint/skeleton prediction can be further divided into: 1) end-to-end estimation models that output 3D keypoint locations given monocular images \cite{pavlakos17volumetric,pavlakos2018ordinal,sun_integralECCV18,Li_ICCV15}; 2) two-stage 2D-to-3D lifting schemes that first estimate 2D human keypoints and then lift to 3D space \cite{Chen_CVPR17,Li_2019_CVPR,martinez_2017_3dbaseline,zhou2019hemlets, su2022virtualpose}. The latter solutions typically outperform end-to-end ones by leveraging superior performance and generalizability of off-the-shelf 2D keypoint detectors. %While keypoint/skeleton human representation is flexible, 
On the other hand, 3D human mesh recovery (HMR) \cite{loper2015smpl}  regresses and outputs mesh parameters, containing richer shape and texture information of the human body. 
Recently, numerous methods \cite{kolotouros2019learning,Arnab_CVPR_2019,bogo2016keep,li2020hybrik} focus on estimating parameters of the Skinned Multi-Person Linear (SMPL)  \cite{loper2015smpl}, a commonly-used volumetric human model with high compatibility, to statistically regress human meshes. 
Several works take steps to leverage a variety of easily-obtained clues, \ie, weak supervision, such as paired 2D landmarks and silhouettes \cite{tan2017indirect, pavlakos2018learning, kanazawa2018end, rong2019delving, Wehrbein_2021_ICCV}.
%For example, Kanazawa \etal \cite{kanazawa2018end} proposed an end-to-end human mesh recovery network, which optimizes network parameters based on 2D keypoints allowing training with in-the-wild images. Kolotouros \etal \cite{kolotouros2019learning}, instead proposed a deep network that combines a regression network with an iterative optimization scheme that fits the SMPL body model to 2D joints in the training loop and then utilized fitted parameters for network supervision for improved human mesh estimation. 
%We refer readers to \cite{zheng2020deep} for an extensive survey of monocular 3D HPE methods.

\noindent \textbf{Multi-view 3D Human Pose Estimation}
%Multi-view 3D pose estimation becomes more robust to depth ambiguities and occlusions compared to single-view solutions. 
Many methods  \cite{dong2019fast,liang2019shape,multiviewpose_ICCV19,rhodin2018unsupervised,pavlakos2017harvesting,zhang2021lightweight} have recently proposed for multi-view 3D HPE. While the majority focuses on 3D body keypoint/skeleton prediction \cite{pavlakos2017harvesting,dong2019fast,rhodin2018unsupervised}, we consider the problem of multi-view 3D HMR, which reconstructs 3D SMPL \cite{loper2015smpl} pose and shape parameters given multiple view images. Existing multi-view HMR methods are all supervised, fusing multi-view with probabilistic modeling \cite{kolotouros2021prohmr} or collaborative learning \cite{li20213d} to regress SMPL parameters.  \citet{liang2019shape} uses additional synthetic image-SMPL pairs to train a multi-view multi-stage regression network. \cite{dong2021shape} aggregates multi-view observation based on  confidence-aware majority voting technique.

%We explicitly aggregate the multi-view knowledge by unprojecting 2D features to canonical 3D space and. progressive regression extensively explores consensus, diversity and balance to deal with the inherent inconsistency among different views.  %Our method demonstrates significant improvement over the other multi-view HMR works \cite{liang2019shape, li20213d} on reconstruction accuracy as we show in Sec.\ref{sec:exp}.
% 1) Our method is geometry-aware, \ie, we extensively utilize camera geometry information in different camera views for cross-view feature learning with mutual enhancement and 3D body reconstruction, that significantly improves reconstruction accuracy as we show in Section \ref{sec:exp}. 
% 2) We utilize synthetically generated multi-view 2D proxy representations, \eg body keypoints or IUVs, instead of images as in \cite{liang2019shape}, as input to the self-supervised HMR network, thus minimizing the synthetic-to-real gap given that generating realistic image scene and background is a much harder and challenging problem. The proposed framework is thus equipped with better generalizability and robustness during inference by leveraging off-the-shelf 2D HPE models with superior performance in in-the-wild scenarios.
% 3) Our proposed method has a two-stage architecture with a coarse-to-fine cross-proxy representation refinement module, that performs sparse (3D body joints) and dense (IUV/body surface) alignment in 3D space across proxy representations iteratively, resulting in improved 3D pose as well as shape recovery accuracy.

% \vspace{-.5em}
\section{Method}
% \vspace{-.5em}
\subsection{Prerequisites}
\label{ssection:pre}
\textbf{3D human mesh parameterization:} Skinned Multi-Person Linear (SMPL) \cite{loper2015smpl} is a parametric model providing independent body shape $\bm{\beta}$ and pose $\bm{\theta}$ parameters with very low-dimensional parameters (\ie, $\bm{\beta} \in \mathbb{R}^{10}$ and $\bm{\theta} \in \mathbb{R}^{72}$). Pose parameters $\bm{\theta} = \{\bm{\theta}_\text{g}, \bm{\theta}_\text{j} \}$ include global body rotation $\bm{\theta}_\text{g} $ (3-DOF) and relative 3D rotations of 23 joints $\bm{\theta}_\text{j} $ (23$\times$3-DOF) in the axis-angle format. Shape parameters include individual heights and weights indicated by the first 10 coefficients of a PCA shape space. 
SMPL provides a differentiable kinematic function $\mathcal{S}$ from $\{\bm{\theta}, \bm{\beta} \}$  to 6,890 mesh vertices: $\bm{v}=\mathcal{S}(\bm{\theta}, \bm{\beta}) \in \mathbb{R}^{6890 \times 3}$. Besides, 3D locations for $N_\text{J}$ joints of interest are obtained as $\bm{j}^\text{3D}=\mathcal{J}\bm{v}$, where $\mathcal{J} \in \mathbb{R}^{N_\text{J} \times 6890}$ is a linear regression matrix.

\textbf{Monocular training data synthesis:} Existing synthetic based HMR methods \cite{sengupta2020synthetic,Sengupta_2021_CVPR,Sengupta_2021_ICCV} generate paired 2D representations (\ie, binary mask, edge and 2D joints) and 3D meshes with SMPL parameters on the fly during training process. At each training step, pose $\{\bm{\theta}_\text{g}, \bm{\theta}_\text{j} \}$ and shape $\bm{\beta}$ are sampled from MoCap \cite{mocap,rogez2016mocap} datasets and prior statistical normal distribution respectively.
%The camera rotation $\bm{R} \in \mathbb{R}^{3 \times 3}$ is set as identity, and 
The camera translation $\bm{T} \in \mathbb{R}^{3}$ is also dynamically sampled from prior distribution. The intrinsic parameters are fixed and represented by focal length $\bm{f} \in \mathbb{R}^{2}$ and image center offset $\bm{t}= [H/2, W/2]$, where $H$, $W$ is rendered image size.
$\{\bm{\theta}_\text{g}, \bm{\theta}_\text{j}, \bm{\beta} \}$ are forwarded into the SMPL model to obtain 3D joints $\bm{j}_\text{3D}$. 2D joints $\bm{j}_\text{2D}$ can be acquired by $\bm{j}_\text{2D} = \bm{f} \mathbf{\Pi}(\bm{j}_\text{3D} +\bm{T})+\bm{t}$, where $\mathbf{\Pi}$ denotes perspective projection. %and $(\cdot)^\text{T}$ indicates transpose. % \bnote{check use of $\bm{f}$ in previous equation -- isn't $\bm{f}$ supposed to be accounted for within $\mathbf{\Pi}$ instead?}. 
% We normalize the $\bm{j}^\text{2D}$ to be from -1 to 1, and denote normalized version as $\bm{j}^\text{2D}$ in the following for simplification.  
The 2D joints $\bm{j}_\text{2D} \in \mathbb{R}^{N_\text{J} \times 2}$ are transformed into 2D Gaussian joint heatmaps $\bm{J} \in \mathbb{R}^{N_\text{J} \times H \times W}$. Another 2D proxy representation, human mask, can be represented by $\bm{M} \in \mathbb{R}^{H \times W}$. The training model utilizes the synthesized paired data with 2D representations $\{\bm{J}, \bm{M}\}$ as input and 3D meshes $\{ \bm{\theta}, \bm{\beta}, \bm{v}, \bm{j}_\text{3D}\}$ as output. 

\textbf{Volumetric triangulation:} Beyond basic algebraic triangulation, volumetric triangulation \cite{iskakov2019learnable} is able to unproject multi-view 2D features along projection rays to fill a shared 3D cube. The cube is a $L \times L \times L$ - sized 3D bounding box in the global space discretized by $G \times G \times G$ volumetric grids, where $G$ represents the number of voxels along each axis. Then each voxel is filled with the global coordinates of the voxel center to get $\bm{V}^{\text{coords}} \in \mathbb{R}^{G\times G\times G\times 3}$. $\bm{V}^{\text{coords}}$ is projected to the image plan to get its corresponding 2D pixel index $\bm{V}^\text{proj} \in \mathbb{R}^{G\times G \times G \times 2}$. Given 2D maps $\bm{F} \in \mathbb{R}^{C\times H \times W}$ in image space, 
we can fill a cube $\bm{V} \in \mathbb{R}^{G \times G\times G \times C}$ by bilinear sampling \cite{jaderberg2015spatial} using  $\bm{V}^\text{proj}$.
The whole process is differentiable and agnostic to the number of views.

% \vspace{-1em}
\subsection{Multi-view Training Data Synthesis}
\label{sec: trdatasyn}
As described above, the synthesis of 2D proxy representations relies on camera extrinsic parameters $\{\bm{R}_{\text{h}\rightarrow \text{c}}, \bm{T}_{\text{h}\rightarrow \text{c}}\}$ (the transformation between human coordinate and camera coordinate) for each camera. 
The intuitive solution of multi-view synthesis is to randomly sample multi-view camera extrinsic parameters around the human body. But the lack of statistical priors makes it not ideal since some views (\eg, from below the human body) can never happen in real scenarios. It also requires manual tuning to ensure as much visible space as possible to avoid large area of blind spot. % will cause ambiguity and training inefficiency.

We first sample one camera setting $\{\bm{R}_{\text{h}\rightarrow \text{c}^1}, \bm{T}_{\text{h}\rightarrow \text{c}^1}\}$ from prior and then extend it to $N$ cameras according to the transformations among cameras in real scenarios. Though there
exists no direct information, \wrt,  relations among camera setups, it can be inferred from the public camera calibrations for multi-view datasets \cite{ionescu2013human3,Trumble:BMVC:2017}. 
Given $\{\bm{R}_{\text{w}\rightarrow \text{c}^n}, \bm{T}_{\text{w}\rightarrow \text{c}^n} \} (n=1...N)$ as the rotation and translation from canonical world coordinate to the $n$-th camera coordinate, we calculate the transformation from the first camera to the other cameras:
\begin{equation}
\label{transcam}
\begin{aligned}
    &\bm{R}_{\text{c}^1\rightarrow \text{c}^n} = \bm{R}_{\text{w}\rightarrow \text{c}^n} \cdot (\bm{R}_{\text{w}\rightarrow \text{c}^1})^\text{T}, \\
    &\bm{T}_{\text{c}^1\rightarrow \text{c}^n}= \bm{T}_{\text{w}\rightarrow \text{c}^n} - \bm{R}_{\text{c}^1\rightarrow \text{c}^n} \cdot \bm{T}_{\text{w}\rightarrow \text{c}^1}~(n=2...N).
\end{aligned}
\end{equation}
Utilizing these priors ensures that our synthetically trained model can better generalize to different testing multi-view images in the wild.

The overall multi-view synthesis process at each training step can be summarized as: (1) sample $\{\bm{\theta}_\text{g}, \bm{\theta}_\text{j}, \bm{\beta}, \bm{T} \}$ following the same strategy as \citet{sengupta2020synthetic}; (2) sample one set of camera settings $\{\bm{R}_{\text{w}\rightarrow \text{c}^n}, \bm{T}_{\text{w}\rightarrow \text{c}^n}|n=1...N \}$ (\eg, $N$  from 2 to 8);  (3) calculate the transformation from camera 1 to the other cameras using Eq.\ref{transcam}; (4) regress and render 2D joints heatmaps and binary mask  $\{\bm{J}^1, \bm{M}^1\}$ under camera-1 using  $\bm{R}_{\text{h}\rightarrow \text{c}^1}$ (from  $\bm{\theta}_\text{g}$) and $\bm{T}_{\text{h}\rightarrow \text{c}^1}$ as camera extrinsic parameters; (5) regress and render 2D representations $\{\bm{J}^n, \bm{M}^n\}$ under other cameras using camera extrinsic parameters below:
\begin{equation}
\label{transh2cam}
\begin{aligned}
    &\bm{R}_{\text{h}\rightarrow \text{c}^n} = \bm{R}_{\text{c}^1\rightarrow \text{c}^n} \cdot \bm{R}_{\text{h}\rightarrow \text{c}^1}, \\
    &\bm{T}_{\text{h}\rightarrow \text{c}^n} = \bm{R}_{\text{h}\rightarrow \text{c}^n} \cdot \bm{T}_{\text{h}\rightarrow \text{c}^1} + \bm{T}_{\text{c}^1\rightarrow \text{c}^n} ~(n=2...N).
\end{aligned}
\end{equation}
We forward $\{\bm{0}, \bm{\theta}_\text{j},\bm{\beta}\}$ into the SMPL statistical model and get $\{\bm{v}, \bm{j}_\text{3D}\}$ in human coordinate system. We further transform $\{\bm{v}, \bm{j}_\text{3D}\}$ with individual camera rotation $\bm{R}_{\text{h}\rightarrow \text{c}^n}$ to $\{\bm{v}^n, \bm{j}_\text{3D}^n\}$ making it specific under each camera rotation. Then we project to 2D joints $\{\bm{J}^n$ and render to  get binary mask $ \bm{M}^n\}$ for each camera $n$.
The training model utilizes the synthesized paired data with 2D representations $\{\bm{J}^n, \bm{M}^n\}$ and 3D meshes $\{\bm{\theta}_\text{j}, \bm{\beta}, \bm{v}^n, \bm{j}_\text{3D}^n\}$ where $\bm{\theta}_\text{j}$ and $\bm{\beta}$ are shared across all views. Note that the aforementioned camera extrinsic $\{\bm{R}_{\text{w}\rightarrow \text{c}^n}, \bm{T}_{\text{w}\rightarrow \text{c}^n}\}$ and intrinsic matrix $\bm{K}$  will also be used in forwarding. % and $\bm{j}_\text{2D}^n  = \bm{f} \mathbf{\Pi}( \bm{j}_\text{3D}^n +\bm{T}_{\text{h}\rightarrow \text{c}^n})+\bm{t}$ will be used as ground-truth for training. 
 
% \vspace{-1em}
\subsection{Learnable Volumetric Calibration}
\label{sec:calibration}
%Compared with the multi-view fusion in one specific 2D image space via algebraic triangulation, we believe multi-view fusion in the shared 3D space can better aggregate dense information. 
The 3D space for fusion is designed to be consistent with the human coordinate rather than world coordinate for mesh regression efficiency and model generalizability. However, for real testing images, we have no access to the transformation from human coordinate to image space which is necessary to do the volumetric triangulation. 
On the other hand, the camera intrinsic parameters $\bm{K}$ are always known and some testing datasets provide transformation from canonical world coordinate to camera coordinate. When this transformation is not available, we can simply define a world coordinate initialized with camera-1.  The transformation between camera-1 and other cameras can be acquired from 2D joints in different views. Please refer to \citet{kocabas2019self} for details. Based on the aforementioned known transformation from canonical world to each camera coordinate (${\bm{R}}_{\text{w}\rightarrow \text{c}^n}$, ${\bm{T}}_{\text{w}\rightarrow \text{c}^n}$), we design volumetric calibration to transform from canonical world to human coordinate ($\hat{\bm{R}}_{\text{w}\rightarrow \text{h}}$, $\hat{\bm{T}}_{\text{w}\rightarrow \text{h}}$) so that multi-view 2D space can be unified in a shared human space.  Initialized with camera-1, the canonical world is first translated to be with the same origin as the human center which can be interpolated \wrt statistical torso length and  2D joints, and then rotated to be consistent with human global pose.  

\textbf{Translation estimation:} %While \cite{iskakov2019learnable} estimates the position of the pelvis through algebraic triangulation, it is not optimal for us \bnote{argument needed}. 
Under each camera view, we obtain the 3D representation of pelvis heatmaps via volumetric triangulation \cite{iskakov2019learnable}. %(we calculate center of 2D hip left and hip right when not accessible) \xnote{when}. 
The per-voxel likelihood for pelvis is obtained by summing up multi-view 3D heatmaps. Via argmax (averaging the 3D positions of the voxels if there are multiple voxels containing the maximum value) we estimate the 3D pelvis position in world space as the translation from world origin to human origin $\hat{\bm{T}}_{\text{w}\rightarrow \text{h}, \text{w}}$ under world space:
\begin{equation}
    \hat{\bm{T}}_{\text{w}\rightarrow \text{h}, \text{w}} = \text{argmax} \sum_n{ \mathcal{V}(\bm{J}_\text{pelvis}^n; {\bm{R}}_{\text{w}\rightarrow \text{c}^n}, {\bm{T}}_{\text{w}\rightarrow \text{c}^n}, \bm{K})},
\end{equation}
% \vspace{-1em}
where $\mathcal{V}(\cdot; \cdot)$ represents volumetric triangulation.
Note we indicate all transformation by uniform sequence with rotation first and then translation, we therefore organize the transformation from human to each camera coordinate:
% \vspace{-1em}
\begin{equation}
\begin{aligned}
    &\hat{\bm{R}}_{\text{h}\rightarrow \text{c}^n} = {\bm{R}}_{\text{w}\rightarrow \text{c}^n} \cdot \hat{\bm{R}}_{\text{w}\rightarrow \text{h}}^\text{T}, \\
    &\hat{\bm{T}}_{\text{h}\rightarrow \text{c}^n} = \bm{T}_{\text{w}\rightarrow \text{c}^n}-\hat{\bm{R}}_{\text{w}\rightarrow \text{c}^n} \cdot \hat{\bm{T}}_{\text{w}\rightarrow \text{h}, \text{w}}. \\
\end{aligned}
\end{equation}
\textbf{Rotation learning:} 
For each camera view $n$, the encoder takes 2D joints heatmaps and binary mask $\{\bm{J}^n, \bm{M}^n \}$ as input and output downsized features $\bm{F}^n \in \mathbb{R}^{C \times H_0 \times W_0}$, where $C$ is channel size.
To learn the rotation from canonical world to human, we unproject $\bm{F}^n$ to the corrected 3D space with human origin and world rotation (acquired by the estimated translation mentioned above).  
We first take average of the unprojected volumetric 3D features among views, then forward the flattened volumetric features into a fully-connected layer $\mathcal{F}_\text{R}$:
\begin{equation}
    \hat{\bm{R}}_{\text{w}\rightarrow \text{h}} = \mathcal{F}_\text{R} (\cfrac{1}{N} \sum_n  \mathcal{V}(\bm{F}^n; \hat{\bm{R}}_{\text{w}\rightarrow \text{c}^n}, \hat{\bm{T}}_{\text{h}\rightarrow \text{c}^n}, \bm{K}, \frac{H}{H_0}, \frac{W}{W_0})%, \frac{H}{H_0}, \frac{W}{W_0}),
\end{equation}
where $\frac{H}{H_0}$, $\frac{W}{W_0}$ are the scale factors from image space to its downsampled feature space.
The output of the fully-connected layer is a continuous 6-dimensional representation \cite{zhou2019continuity} which can be converted to a discontinuous Euler rotation matrix  $\hat{\bm{R}}_{\text{w}\rightarrow \text{h}}$. 
Note that both volumetric triangulation (grid sampling) and transformation are differentiable, thus $\hat{\bm{R}}_{\text{w}\rightarrow \text{h}}$ can be learned via projecting the predicted 3D mesh to 2D.

We note that the aforementioned calibration from canonical world to human is to unify diverse camera setups in common human space for efficient multi-view learning. The currently available $\bm{K}$ and $\hat{\bm{T}}_{\text{h}\rightarrow \text{c}^n}$ is not learnable and only a rough estimation which aims to correct the 3D cube center to be around the pelvis center. For more accurate joint learning together with the predicted 3D mesh, we employ reprojection loss with orthographic projection following \citet{kanazawa2018end}. Specifically, we learn camera parameters from a fully-connected layer $\mathcal{F}_\text{OP}$ for individual view: $\hat{\bm{\pi}}^n = \mathcal{F}_\text{OP}(\bm{F}^n)$, where $\hat{\bm{\pi}}^n = [\hat{\pi}_\text{s}^n, \hat{\bm{\pi}}_\text{t}^n]$, $\hat{\pi}_\text{s}^n \in \mathbb{R}$ is the scale factor and $\hat{\bm{\pi}}_\text{t}^n \in \mathbb{R}^2$ is translation. %Note $\hat{\bm{\pi}}$ is different across views. 

% \vspace{-1em}
\subsection{Progressive Multi-view Aggregation}
From the aforementioned volumetric calibration, we are able to obtain the volumetric features $\bm{V}_\text{F}^n \in \mathbb{R}^{G \times G \times G \times C}$ for each camera-$n$ in the uniform 3D human space:
\begin{equation}
\bm{V}_\text{F}^n = \mathcal{V}(\bm{F}^n; \hat{\bm{R}}_{\text{h}\rightarrow \text{c}^n}, \hat{\bm{T}}_{\text{h}\rightarrow \text{c}^n}, \bm{K}, \frac{H}{H_0}, \frac{W}{W_0}).
\end{equation}
$\bm{V}_\text{F}^n$ from multiple views are fused (averaged or weighted summation), flattened, and then passed to the regressor $\mathcal{R}$ to predict pose and shape parameters $\bm{\Theta} = \{ \hat{\bm{\theta}_\text{j}}, \hat{\bm{\beta}} \}$. Note that we only predict 23 joints rotation $\hat{\bm{\theta}_\text{j}}$ shared by all views (excluding global root orientation) as we have learnt view-specific global orientation $\hat{\bm{R}}_{\text{h}\rightarrow \text{c}^n}$ in volumetric calibration.
%Sec.\ref{sec:calibration}. 

Following the standard iterative error feedback (IEF) procedure \cite{kanazawa2018end}, we employ three iterative regressions to optimize $\bm{\Theta}$ and further propose progressive multi-view aggregation. %including consensus learning, diversity learning, and balance learning. 
We introduce how we progressively learn from consensus/diversity information from intersection and union occupancy mask, and then how we balance the multi-view 3D features from the consistency between consensus and diversity preserving mesh prediction. 

\subsubsection{Consensus and diversity sampling:} Under each camera-$n$, we consider all the possible nonzero areas of 2D joints heatmaps $\bm{J}^n$ and binary mask $\bm{M}^n$ as 2D occupancy mask $\bm{O}^n \in \{ 0, 1\}^{H\times W}$ and then we obtain the volumetric occupancy mask in 3D human space $\bm{V}_\text{O}^n \in \{ 0, 1\}^{G \times G \times G}$: 
\begin{equation}
\begin{aligned}
&\bm{O}^n = \mathbf{1}((\sum\nolimits_{j=1}^{N_\text{j}}{\bm{J}_j^n}+\bm{M}^n)>0), \\
&\bm{V}_\text{O}^n = \mathcal{V}(\bm{O}^n; \hat{\bm{R}}_{\text{h}\rightarrow \text{c}^n}, \hat{\bm{T}}_{\text{h}\rightarrow \text{c}^n}, \bm{K}).\\
\end{aligned}
\end{equation}
To efficiently aggregate the volumetric occupancy masks, we take intersection of $\{\bm{V}_\text{O}^n| n=1...N\}$ as $\bm{V}_\text{O}^\text{I}$ representing the area of interest shared by all views (consensus), and take union of them as $\bm{V}_\text{O}^\text{U}$ which representing the area of interest masked by at least one view (diversity):
\begin{equation}
\bm{V}_\text{O}^\text{I} =  \text{min}_n{\bm{V}_\text{O}^n} ,\\
~\bm{V}_\text{O}^\text{U} =  \text{max}_n{\bm{V}_\text{O}^n}, \\
\end{equation}
where $\bm{V}_\text{O}^\text{I}, \bm{V}_\text{O}^\text{U} \in \{0,1\}^{G \times G \times G}$.
To achieve consensus while also maintaining the diversity that is inherent among the multiple views, we mask the 3D volumetric features  $\{\bm{V}_\text{F}^n\}$ spatial-wisely with occupancy intersection $\bm{V}_\text{O}^\text{I}$ and occupancy union $\bm{V}_\text{O}^\text{U}$ respectively:
\begin{equation}
\bm{V}_\text{F}^\text{I} = \bm{V}_\text{O}^\text{I}\odot {\frac{1}{N} \sum_n\bm{V}_\text{F}^n},\\
~ \bm{V}_\text{F}^\text{U} = \bm{V}_\text{O}^\text{U}\odot {\frac{1}{N} \sum_n\bm{V}_\text{F}^n}. \\
\end{equation}
where $\odot$ is Hadamard product, and $\bm{V}_\text{F}^\text{I}, \bm{V}_\text{F}^\text{U} \in \mathbb{R}^{G \times G \times G \times C}$ indicate the fused 3D features in consensus and diversity occupancy area respectively. Note that the diversity occupancy area is introduced to tolerate possible bias of one-view 3D occupancy which is easily caused by inaccurate camera calibration. We progressively forward $\bm{V}_\text{F}^\text{I}, \bm{V}_\text{F}^\text{U}$ into the regressor $\mathcal{R}$ so that the regressor can first focus on the features in area commonly occupant by all views ($\bm{V}_\text{O}^\text{I}$), and then consider the features in all possible occupant areas ($\bm{V}_\text{O}^\text{U}$). The output of $\mathcal{R}$ can be represented by $\bm{\Theta}_1$ and $\bm{\Theta}_2$:
\begin{equation}
    \bm{\Theta}_1 = \bm{\Theta}_0 + \mathcal{R}(\bm{V}_\text{F}^\text{I} ;\bm{\Theta}_0), \\
    \bm{\Theta}_2 = \bm{\Theta}_1 + \mathcal{R}(\bm{V}_\text{F}^\text{U} ;\bm{\Theta}_1), \\
\end{equation}
where $\bm{\Theta}_0$ is reposed pose and mean shape for initialization. 

\subsubsection{Multi-view balance via consistency weighting: } 
\label{subsubsec:consistency}
We take average of the multi-view 3D features for intersection and union fusion in the first two regression iteration. % as we lack reliable clue among the views , and 
At the last iteration of regression we utilize the current 3D mesh prediction as evidence to seek consistency among views. Specifically, we project the 3D mesh to individual 2D image spaces for spatial-wise consistency as fusion confidence under each camera. 
Given the output of the regressor after the second iteration $\bm{\Theta}_2 = \{ \hat{\bm{\theta}}_\text{j}, \hat{\bm{\beta}}\}$,  SMPL takes $\{ \bm{0}, \hat{\bm{\theta}}_\text{j}, \hat{\bm{\beta}} \}$ to infer 3D vertices $\hat{\bm{v}}$ and 3D joints $\hat{\bm{j}}^\text{3D}$. 
Using a differentiable renderer \cite{ravi2020pytorch3d}, we generate a body mask $\hat{\bm{M}}^n$ from $\hat{\bm{v}}$ according to the camera parameters $\{\hat{\bm{R}}_{\text{h}\rightarrow \text{c}^n}, \hat{\bm{\pi}}^n \}$. 
We also obtain the reprojected 2D joints $\hat{\bm{j}}_\text{2D}^n = \hat{{\pi}}_s^n \cdot \mathbf{\Pi}(\hat{\bm{R}}_{\text{h}\rightarrow \text{c}^n} \cdot \hat{\bm{j}}_\text{3D}^\text{T}) + \hat{\bm{\pi}}_t^n$ and convert to heatmap version  $\hat{\bm{J}}^n$ for each camera-$n$. Comparing these reprojected 2D representations $\{\hat{\bm{J}}^n, \hat{\bm{M}}^n \}$  with the input $\{{\bm{J}}^n, {\bm{M}}^n \}$, we calculate the consistency map $\bm{\phi}^n \in \mathbb{R}^{H \times W}$ under each 2D image space: 
\begin{equation}
    \bm{\phi}^n = \frac{1} {\epsilon + |\hat{\bm{M}}^n - \bm{M}^n| + \sum_{j=1}^{N_\text{j}} {|\hat{\bm{J}}_j^n - \bm{J}_j^n|/N_\text{j}}}. \\
\end{equation}
We further unproject $\bm{\phi}^n$ to get the volumetric consistency representation $\bm{V}_{\phi}^n \in \mathbb{R}^{G \times G \times G} $ of each view under the commonly shared 3D human space, and then normalize these volumetric consistency among $N$ camera views: 
\begin{equation}
    \bar{\bm{V}}_{\phi}^n = \frac{\bm{V}_{\phi}^n}{\sum_n \bm{V}_{\phi}^n} = \frac{\mathcal{V}(\bm{\phi}^n; \hat{\bm{R}}_{\text{h}\rightarrow \text{c}^n}, \hat{\bm{T}}_{\text{h}\rightarrow \text{c}^n}, \bm{K})}{\sum_n \mathcal{V}(\bm{\phi}^n; \hat{\bm{R}}_{\text{h}\rightarrow \text{c}^n}, \hat{\bm{T}}_{\text{h}\rightarrow \text{c}^n}, \bm{K})}.
\end{equation}
Taking $\bar{\bm{V}}_{\phi}^n$ as view-specific volumetric confidence, we are able to balance volumetric 3D features ${\bm{V}}_\text{F}^n$ under each camera into ${\bm{V}}_\text{F}^\text{B}$ where the view less consistent with jointly reached  3D mesh is given less confidence for fusion:
\begin{equation}
{\bm{V}}_\text{F}^\text{B} = \sum_n \bar{\bm{V}}_{\phi}^n \odot {\bm{V}}_\text{F}^n.
\end{equation}
At the final iteration, the regressor takes the consistency balanced 3D feature ${\bm{V}}_\text{F}^\text{B}$ for the final prediction :$\bm{\Theta}_3 = \bm{\Theta}_2 + \mathcal{R}(\bm{V}_\text{F}^\text{B} ;\bm{\Theta}_2).$

\subsection{Loss Function}
\label{sec:loss}
%As described in Section \ref{subsubsec:consistency} from 
As described above, from the final description $\bm{\Theta}_3 = \{ \hat{\bm{\theta}}_\text{j}, \hat{\bm{\beta}}\}$ we have vertices $\hat{\bm{v}}$ and 3D joints $\hat{\bm{j}}_\text{3D}$ based on SMPL regression. With camera parameters $\{ \hat{\bm{R}}_{\text{h}\rightarrow \text{c}^n}, \hat{\bm{\pi}}^n\}$, we infer to vertices and 3D joints under each camera rotation: $\hat{\bm{v}}^n =  \hat{\bm{R}}_{\text{h}\rightarrow \text{c}^n} \cdot \hat{\bm{v}}^\text{T}$, $\hat{\bm{j}}_\text{3D}^n =  \hat{\bm{R}}_{\text{h}\rightarrow \text{c}^n} \cdot \hat{\bm{j}}_\text{3D}^\text{T}$. $\hat{\bm{j}}_\text{3D}^n$ is then projected to 2D joints $\hat{\bm{j}}_\text{2D}^n$ with orthographic projection.
%We have prediction and supervision in terms of vertices, 2D joints, 3D joints and SMPL parameters:
The overall loss for mesh regression is therefore defined as
\begin{equation}
\label{eq:loss}
\begin{aligned}
&\Loss ( \hat{\bm{R}}_{\text{h} \rightarrow \text{c}^n},  \hat{\bm{\theta}}_\text{j}, \hat{\bm{\beta}}, \hat{\bm{v}}^n, \hat{\bm{j}}_\text{2D}^n, \hat{\bm{j}}_\text{3D}^n, {\bm{R}}_{\text{h} \rightarrow \text{c}^n},  {\bm{\theta}}_\text{j}, {\bm{\beta}}, , \bm{v}^n, {\bm{j}}_\text{2D}^n, {\bm{j}}_\text{3D}^n) \\
=& {\omega_\theta} {\Loss_2( \hat{\bm{\theta}}_\text{j}, {\bm{\theta}}_\text{j}}) + {\omega_\beta} {\Loss_2( \hat{\bm{\beta}}, {\bm{\beta}})} + \sum_{n=1}^N{\omega_\text{R}}\Loss_2( \hat{\bm{R}}_{\text{h} \rightarrow \text{c}^n}, {\bm{R}}_{\text{h} \rightarrow \text{c}^n})  \\
& \omega_\text{v}\Loss_2 (\hat{\bm{v}}^n, \bm{v}^n) +\omega_\text{j2D}\Loss_2( \hat{\bm{j}}_\text{2D}^n, {\bm{j}}_\text{2D}^n) + {\omega_\text{j3D}}\Loss_2( \hat{\bm{j}}_\text{3D}^n, {\bm{j}}_\text{3D}^n) \\
\end{aligned}
\end{equation}
where $\Loss_2$ denotes the mean square error (MSE), and $\omega_{\theta}$, $\omega_{\beta}$, $ {\omega_\text{R}}$ $\omega_\text{v}$, $\omega_\text{j2D}$, and $\omega_\text{j3D}$ indicate weights for joints pose and shape, view-specific global pose, vertices, 2D joints, 3D joints respectively. Note for $\Loss_2( \hat{\bm{\theta}}_\text{j}, {\bm{\theta}}_\text{j})$ both are first converted to rotation matrix for the MSE loss.

% \vspace{-10mm}

\begin{table}[]
\centering
% \resizebox{\columnwidth}{!}
\scriptsize
{
\begin{tabular}{cccccc}
\hline
\multirow{3}{*}{Method}  &\multicolumn{3}{c}{Training Requirement} &\multicolumn{2}{c}{Metrics}  \\ \cmidrule(lr){2-6}
&\parbox[t]{5mm}{\rotatebox[origin=c]{60}{\shortstack[c]{Superv.}}}    &\parbox[t]{5mm}{\rotatebox[origin=c]{60}{\shortstack[c]{Multi-view \\ Imagery}}}
&\parbox[t]{5mm}{\rotatebox[origin=c]{60}{\shortstack[c]{Temporal \\ Sequence}}}  
&{MPJPE$\downarrow$} &{PMPJPE$\downarrow$} \\\hline
Rhodin \etal \shortcite{rhodin2018unsupervised} &J3D &\cmark  &\xmark  &131.7 &98.2 \\
% Remelli \etal \cite{remelli2020lightweight} &N &\cmark & &\cmark &calibration &30.2 & \\
% MetaFuse \cite{xie2020metafuse} &N &\cmark & &\cmark &- &29.3 &-\\
PVH-TSP \shortcite{Trumble:BMVC:2017} &J3D  &\cmark &\cmark   &87.3 &- \\
Tome \etal \shortcite{tome2018rethinking} &J3D  &\cmark  &\xmark  &52.8 &- \\
Remelli \etal \shortcite{remelli2020lightweight} &J3D  &\cmark  &\xmark &30.2 &-\\
Bartol \etal \shortcite{Bartol_2022_CVPR} &J3D  &\cmark  &\xmark  &29.1 &-\\
Pavlakos \etal \shortcite{pavlakos2017harvesting} &\xmark &\cmark &\xmark  &56.9 &-\\
% Chen \etal \shortcite{chen2019unsupervised} &\xmark &\xmark &\xmark &\xmark &55 &-\\
% Chen \etal \shortcite{chen2019unsupervised} &\xmark &\xmark &\cmark &\xmark &51 &-\\
\hline
Trumble \etal \shortcite{trumble2018deep}$^*$ &J3D  &\cmark &\cmark &62.5 &- \\
Liang \etal \shortcite{liang2019shape}$^*$ &Mesh  &\cmark &\xmark  &79.8 &45.1 \\
Li \etal \shortcite{li20213d}$^*$ &Mesh  &\cmark &\xmark &64.8 &43.8 \\
ProHMR \shortcite{kolotouros2021prohmr}$^*$ &Mesh  &\xmark &\xmark &62.2 & \textbf{34.5}\\
Ours$^*$ &\xmark &\xmark &\xmark  &\textbf{53.8} &42.4\\
\hline
\end{tabular}}
% \vspace{-1em}
\caption{Comparisons on MPJPE and PMPJPE (both in mm) on the Human3.6M test sets with multi-view human pose/mesh estimation methods. $^*$ indicates method which can recover human shape beyond human pose. 
}
\label{tab:h36m}
%\vspace{-1em}
\end{table}

\section{Experiments}
\label{sec:exp}
Note that we train only one model generalized to different test sets, whereas most counterparts train individual model, with corresponding training data, for each specific test set.

\subsection{Datasets And Metrics}
\textbf{Training data.}
    Following the protocol by \citet{sengupta2020synthetic}, we sample SMPL pose parameters from the training sets of UP-3D \cite{lassner2017unite}, 3DPW \cite{von2018recovering}, and the five training subjects (S1, S5, S6, S7, S8) of Human3.6M \cite{ionescu2013human3}. %Note that each time we only sample one SMPL pose for multiview synthesis and the camera setup are sampled based on prior knowledge. 
%The sampling of shape parameters follows the procedure of prior work \cite{sengupta2020synthetic}. 

\noindent\textbf{Evaluation data.}
To evaluate the generalizability of our method, we test on both indoor and outdoor datasets with different number of views. Human3.6M is one of the largest/most commonly used 3D human pose estimation benchmark with SMPL annotation. Besides, we evaluate on TotalCapture (in-the-studio) and SkiPose (in-the-wild). Both only have 3D joint annotations (no body shape ones). 

\textbf{Human3.6M:}   
The Human3.6M dataset \cite{ionescu2013human3} provides a total of 3.6 million frames in  synchronized four-views.
The camera placement is slightly different for each of the seven subjects. We follow the most popular protocol 1, testing on subjects S9, S11. We report mean per joint position error (MPJPE) and mean per joint position error after rigid alignment with Procrustes analysis (PMPJPE) on the 17 joints in H3.6M definition.  %Besides, we evaluate per-vertex error (PVE) and per-vertex error after rigid alignment with Procrustes analysis (PPVE) as shape metrics.

\textbf{TotalCapture:} 
TotalCapture dataset \cite{Trumble:BMVC:2017} consists of 1.9
million frames, captured from 8 calibrated full HD video
cameras recording at 60Hz. Following the typical data split \cite{Trumble:BMVC:2017}, we use “Walking2”, “Freestyle3”, and “Acting3” on subjects 1, 2, 3, 4, 5 for testing. We report the mean per joint position error (MPJPE) as 3D pose metric for comparison with prior arts. 

\textbf{Ski-Pose PTZ:} 
This dataset \cite{sporri2016reasearch,rhodin2018learning} records competitive alpine skiers performing giant slalom runs with eight moving cameras. The cameras are rotating and zooming to keep the alpine skier in the field of view. We follow the typically used metrics: MPJPE, PMPJPE, and percentage of correct keypoints (PCK) thresholded at 150mm \cite{mehta2017monocular}. %, and the area under the curve (AUC) over a range of PCK thresholds
% \vspace{-2mm}

% \begin{table}[]
% \centering
% \resizebox{\columnwidth}{!}
% {
% \begin{tabular}{cccccc}
% \hline
% Method &Temporal Prior &MPJPE$\downarrow$ &PMPJPE$\downarrow$ &PVE$\downarrow$ &PPVE$\downarrow$\\\hline
% STRAP \shortcite{sengupta2020synthetic} &\xmark &87.0 &59.3 &117.3 &82.1  \\
% Skeleton2Mesh \shortcite{yu2021skeleton2mesh} &\xmark &87.1 &55.4 &120.8 &80.8\\
% Skeleton2Mesh \shortcite{yu2021skeleton2mesh} &\textit{Partial/Whole}
% &- &- &- &59.7/51.0\\\hline % 
% % Ours (1 camera) &\xmark &84.3 &57.1 &109.7 &74.2\\ 
% Ours (2 cameras) &\xmark &59.1 &45.0 &97.9 &60.2\\ 
% Ours (3 cameras) &\xmark &55.3 &43.1 &93.7 &55.1 \\
% Ours (4 cameras) &\xmark  &53.8 &42.4 &89.7 &52.3\\
% \hline
% \end{tabular}}
% % \vspace{-1em}
% \caption{Comparisons with synthetic-based monocular human mesh recovery methods on the Human3.6M dataset on MPJPE, PMJPE as pose metrics and PVE, PPVE as shape metrics (all in mm). `\textit{Partial/Whole}' denotes utilizing discontinuous frames and whole sequences as temporal priors. }
% \label{tab:h36mab1}
% %\vspace{-1em}
% \end{table}

% \vspace{-5mm}
\begin{table*}[]
\centering
% \resizebox{\columnwidth}{!}
\scriptsize
{
\begin{tabular}{cc|ccccccc}
\hline
\multirow{2}{*}{Superv.} &\multirow{2}{*}{Method} &\multicolumn{3}{c}{Subjects (S1,2,3)} &\multicolumn{3}{c}{Subjects (S4,5)} &\multirow{2}{*}{Mean} 
 \\
% & &\parbox[t]{10mm}{\rotatebox[origin=c]{15}{\shortstack[c]{Walking2}}} &\parbox[t]{10mm}{\rotatebox[origin=c]{15}{\shortstack[c]{Action3}}} &\parbox[t]{12mm}{\rotatebox[origin=c]{15}{\shortstack[c]{Freestyle3}}} &\parbox[t]{10mm}{\rotatebox[origin=c]{15}{\shortstack[c]{Walking2}}} &\parbox[t]{10mm}{\rotatebox[origin=c]{15}{\shortstack[c]{Action3}}} &\parbox[t]{12mm}{\rotatebox[origin=c]{15}{\shortstack[c]{Freestyle3}}}
& &Walking2 &Acting3 &Freestyle3 &Walking2 &Acting3 &Freestyle3
\\\hline
\multirow{3}{*}{{Image-Joints3D pairs} (\textit{In})} &PVH \shortcite{Trumble:BMVC:2017} &48.3 &94.3 &122.3 &84.3 &154.5 &168.5 &107.3 \\
&Tri-CPM \shortcite{wei2016convolutional} &79.0 &106.5 &112.1 &79.0 &73.7 &149.3 &99.8 \\
&IMUPVH \shortcite{Trumble:BMVC:2017} &30.0 &49.0 &90.6 &36.0 &109.2 &112.1 &70.9 \\\hline
{Image-Joints3D pairs} (\textit{In}) &Trumble \etal \shortcite{trumble2018deep}$^*$ &42.0 &59.8 &120.5 &58.4 &103.4 &162.1 &85.4\\
Image-Mesh pairs (\textit{Cr}) &ProHMR \shortcite{kolotouros2021prohmr}$^*$ &125.7 &118.9 &134.3 &131.9 &125.2 &135.8 &127.8\\
Self &Ours$^*$ &66.1 &69.3 &58.9 &64.4 &79.1 &61.3 &\textbf{64.2} \\
\hline
\end{tabular}}
\vspace{-1em}
\caption{Comparison of multi-view 3D human pose/mesh estimation methods in terms of 3D pose errors MPJPE $\downarrow$ (mm) on TotalCapture test set. $^*$ indicates method which can recover both body pose and shape. `\textit{In}' denotes training data from the same dataset as the testing data. `\textit{Cr}' denotes training data is from cross/different dataset. We report the results of ProHMR trained with Human3.6M and additional 2D keypoints fitting. % during testing, while ours only uses 2D keypoints as input without 2D fitting during testing. 
}
\label{tab:totalcapture}
% \vspace{-1em}
\end{table*}
% \vspace{-5mm}
\subsection{Implementation Details}
\noindent\textbf{Synthetic training.}
We generate multi-view paired data on the fly. To simulate noise and discrepancy between 2D joints and mask prediction and among different views, we apply a series of processing and augmentations.
%including randomly masking one of the six body parts  (same as PartDrop in \cite{zhang2020learning}),  randomly masking one of the six body parts (head, torso, left/right arm,  left/right leg) on IUV map, randomly occluding the IUV map with a dynamically-sized rectangle, and randomly perturbing the 2D joints position. All hyper-parameters can be .
Training is done using Adam \cite{kingma2014adam} optimizer for 6 epochs with a learning rate of $1e^{-4}$ and a batch size of 16. It takes \textapprox3 days on one A100 GPU.
% We use Adam \cite{kingma2014adam} optimizer to train for 6 epochs with a learning rate of $1e^{-4}$ and a batch size of 16. The training takes around 3 days on one A100 GPU. 

\noindent\textbf{Testing.} We infer 2D joints on the testing images with the pretrained Keypoint-RCNN \cite{he2017mask} with ResNet-50 backbone. We predict the human mask using pretrained DensePose-RCNN \cite{he2017mask} with ResNet-101 backbone. %Since 3DPW test images may have multiple persons, we use the same protocol as prior work \cite{kolotouros2019learning} to get the bounding box for the target person by using the scale and center information and get the 2D representations with maximum IOU with the target bounding box. 
For consistency with training, we crop both the masks and 2D joints heatmaps with a scale of 1.2 before forwarding them to the network for 3D mesh inference .

\subsection{Results}
\textbf{Human3.6M.} Table \ref{tab:h36m} compares our method on test sets of Human3.6M Protocol 1 with other multi-view human pose estimation methods. The counterparts utilize fully-paired 3D annotation or auxiliary clues, \eg, multi-view images. Compared with the other self-supervised method \cite{pavlakos2017harvesting}, our method 1) is able to predict human shape beyond human pose, 2) does not rely on any auxiliary requirement. 
Comparing to the pose-only methods (top half), we note a large performance gap between self-supervised methods and fully-supervised arts. But our self-supervised mesh recovery method is comparable to the fully-supervised mesh recovery SOTA method \cite{kolotouros2021prohmr}. 
%It is also worth noting that our self-supervised method achieves better pose estimation performance (on both MPJPE and PMPJPE) than prior arts of fully supervised multi-view human mesh recovery methods \cite{liang2019shape, li20213d}. 
%With additional utility of body shape estimation, our synthetic-data-driven method is still competitive when compared with those fully-supervised methods which are only able to predict human pose. 
% Table \ref{tab:h36mab1} compares our method with synthetic-based monocular HMR methods on pose and shape metrics. The results show that we efficiently aggregate the multi-view representations in 3D space to compensate for ambiguity in single view. 

\begin{table}[t]
\centering
\resizebox{\columnwidth}{!}
% \scriptsize
{
\begin{tabular}{cc|cc|ccc}
\hline
% \multirow{2}{*}{Method} &Recover &\multirow{2}{*}{Supervision} &\multicolumn{2}{c|}{Images Requirement} &\multirow{2}{*}{MPJPE$\downarrow$} &\multirow{2}{*}{PMPJPE$\downarrow$} &\multirow{2}{*}{PCK$\uparrow$} \\
% &Body Shape & &Training &Testing \\
Superv. &Method &Train  &Test &MPJPE$\downarrow$ &PMPJPE$\downarrow$ &PCK$\uparrow$ \\
\hline
Semi &Pavllo \etal \shortcite{pavllo20193d} &\textit{T} &\textit{T} &106.0 &88.1 &- \\
\multirow{2}{*}{Weak} &AdaptPose \shortcite{gholami2021adaptpose}&\textit{T} &\textit{S} &99.4 &83.0 &-\\
&Rhodin \etal \shortcite{rhodin2018learning} &\textit{MV}  &\textit{S} &85.0 &- &- \\ 
Self &CanonPose \shortcite{wandt2021canonpose} &\textit{MV} &\textit{S} &128.1 &89.6 &67.1 \\\hline
\multirow{2}{*}{Full} &ProHMR \shortcite{kolotouros2021prohmr}$^*$ &\textit{S(Cr)} &\textit{S} &122.7 &82.6 &73.4\\
&ProHMR \shortcite{kolotouros2021prohmr}$^*$ &\textit{S(Cr)}  &\textit{MV} 
&105.7 &73.1 &80.3 \\
%&192.6 &101.5 &51.4\\
\multirow{2}{*}{Self} &Ours$^*$&\xmark &\textit{S} &109.2 &72.6 &77.5\\
&Ours$^*$&\xmark &\textit{MV} &\textbf{89.6} &\textbf{64.8} &\textbf{86.0}  \\
\hline
\end{tabular}}
\vspace{-1em}
\caption{Comparison with 3D pose/mesh estimation methods on Ski-Pose PTZ test set in terms of MPJPE, PMPJPE (in mm) and PCK (\%).  $^*$ indicates additional ability to recover body shape. `\textit{S}', `\textit{T}', `\textit{MV}'  denotes single, temporal consistent, and multi-view images respectively. `\textit{Cr}' denotes cross dataset (Human3.6M) different with the test data. }
\label{tab:skipose}
\end{table}

\noindent\textbf{TotalCapture.} Table \ref{tab:totalcapture} compares our HMR method with other multi-view human pose/mesh estimation methods on TotalCapture test set. All the methods for comparison take all view images as input for inference. We note that our self-supervised human pose and shape estimation method demonstrates better performance in terms of MPJPE (mm) than the fully-supervised methods \cite{Trumble:BMVC:2017,wei2016convolutional,trumble2018deep} which are only able to predict human pose. The superiority of our method beyond the others is more obvious on S4,5 which are unseen subjects in TotalCapture training set, as those methods trained with paired image-3D annotation can hardly generalize. We can see the SOTA multi-view mesh predication method \cite{kolotouros2021prohmr} (trained with Human3.6M) can hardly generalize to TotalCapture test set though they are both in-the-studio datasets.
In contrast, our model trained with synthetic 2D-3D pairs better generalize to diverse unseen data.

\noindent\textbf{SkiPose PTZ.} 
The comparison on the test set of SkiPose is shown in Table \ref{tab:skipose}.
%other methods \cite{pavllo20193d,gholami2021adaptpose,rhodin2018learning} heavily rely on supervision and related temporal or multi-view images for training. 
%In contrast, our method has no requirements of training images/annotations and, at the same time, achieves better results on all the three metrics. 
%We also compare with the SOTA  self-supervised single view human pose estimation method \cite{wandt2021canonpose}. 
With additional utility of body shape estimation, our method still outperforms self-supervised SOTA human pose estimation method \cite{wandt2021canonpose}. Comparison with SOTA supervised multi-view human mesh recovery method (ProHMR \cite{kolotouros2021prohmr} trained with Human3.6M) shows our purely synthetic-data-trained method has superior generalization to in-the-wild scenarios even when ProHMR uses 2D keypoints fitting during testing while ours only use 2D keypoints as input.

%The proposed method can deal with all these difficulties since it does not require a calibrated or static setup and works with multiple synchronised cameras

\begin{figure}[t]
	\centering
	%\vspace{-1em}
	\includegraphics[width=1\linewidth, height=6cm]{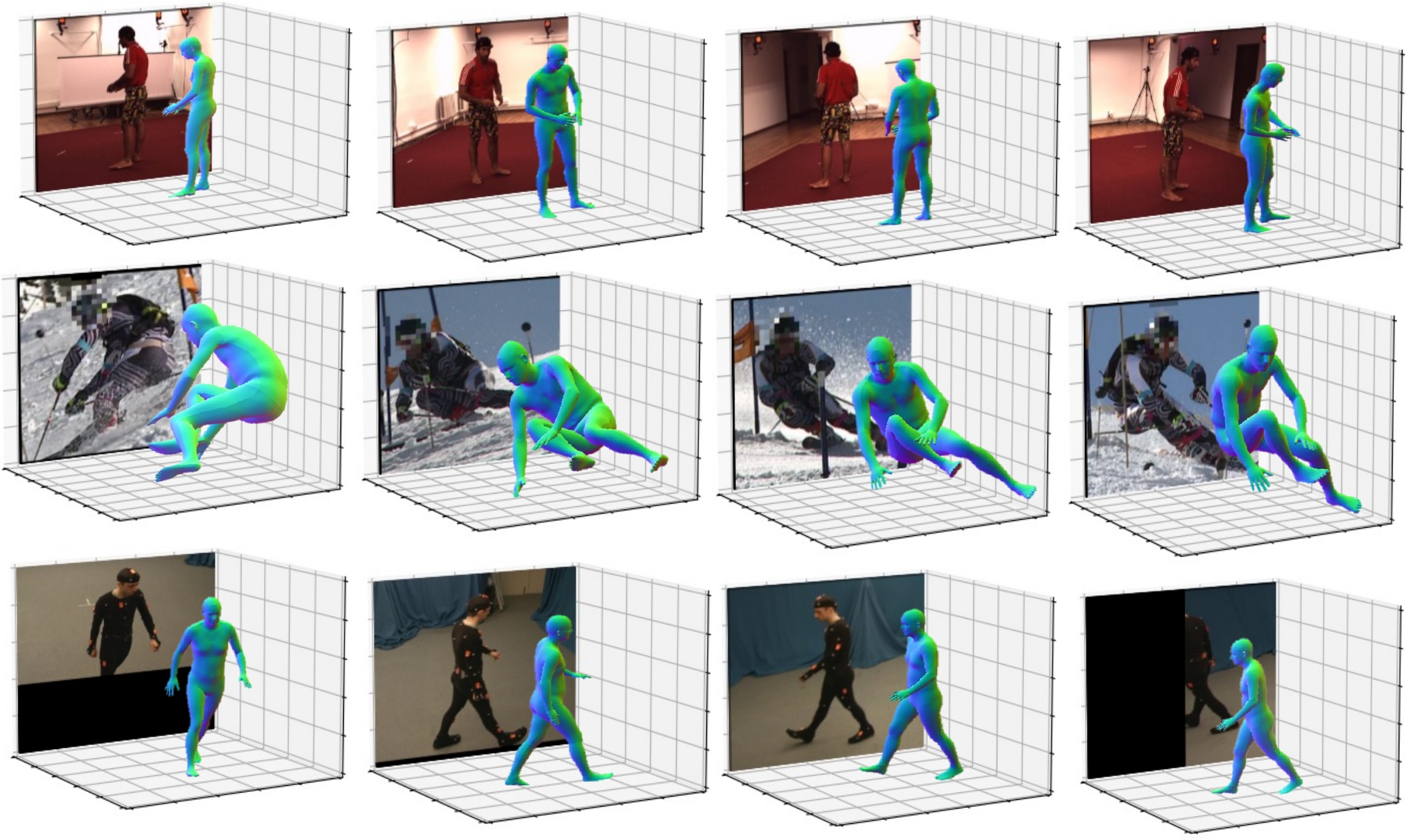}
	%\vspace{-1em}
	\caption{Visualization of reconstructed SMPL mesh for Human3.6M (row 1), SkiPose (rows 2) and TotalCapture (row 3) datasets (select four views for latter two datasets). %We only select four views for visualization from the SkiPose and TotalCapture.  \xnote{edit}
	}
	\label{fig:vismesh}
	\vspace{-1em}
\end{figure}

\noindent\textbf{Qualitative results.}
Figure \ref{fig:vismesh} gives qualitative examples where we visualize our predicted 3D mesh on images from the three testing datasets. Notably, we train only one model and test on different datasets. The results demonstrate the robustness and generalization ability of our synthetically-trained model to various unseen in-the-wild data.

% \noindent
% More quantitative and qualitative results can be found in supplementary materials.

\section{Conclusion}
To the best of our knowledge, we propose the first multi-view human mesh recovery method based on self-supervised synthetic training. 
% We propose a novel multi-view human mesh recovery method based on self-supervised synthetic training, to the best of our knowledge the first self-supervised multi-view human mesh recovery algorithm. 
Our solution first extracts intermediate 2D representations from each view and projects the corresponding features to a 3D canonical space with learnable volumetric calibration.
% 2D representations are extracted from intermediate representations and projected to 3D canonical space with learnable volumetric feature triangulation.
Multi-stage progressive regressors then iteratively refine estimated mesh parameters based on different feature-sampling criteria. Extensive evaluations demonstrate the efficacy and superior performance of the proposed method, especially in challenging in-the-wild scenarios where 1) single-view-based methods suffer from depth ambiguities and 2) supervision-based methods have no access to any prior of in-domain image and annotation. 

% \clearpage
% \bibliographystyle{splncs04}nobibliography
% \nobibliography{egbib}
\bibliography{egbib}

\end{document}

\maketitle

\appendix
\beginsupplement

%%%%%%%%% BODY TEXT
This document provides material supplementing the main manuscript. We first detail the data synthesis and network design during training. Then we share additional ablation studies for pipeline design,
as well as different model structures for off-the-shelf detectors. 
Section \ref{sec:vis} shows more qualitative results with comparisons between the baseline method and the proposed method with progressive regression.

\section{Training Details}

Paired synthetic data (2D joints, silhouettes, and SMPL mesh) are generated on the fly to train the multi-view regression network from intermediate representations to human mesh. The whole synthesis process can be divided into data sampling, data generation, and data augmentation. 
Details are provided below, and the values of related hyper-parameters are shared in Table \ref{tab:aug}.

\textbf{ Data sampling.} 
At each training step, we independently sample SMPL (\ie, pose and shape) parameters and camera parameters.
Pose parameters $\bm{\theta}$ are sampled from MoCap priors of Human3.6M \cite{ionescu2013human3} training set, UP-3D \cite{lassner2017unite} training set, and 3DPW \cite{von2018recovering} training set.  Shape parameters $\bm{\beta}$ are sampled from independent normal distribution $\beta_n \sim \mathcal{N}(\mu_n, \sigma_n^2) (n=1,\ldots,10)$.
For camera extrinsic parameters, we sample the camera translation from the same distribution as \cite{sengupta2020synthetic}. Note that the camera translation for different views are sampled independently. We use the root orientation from Mocap as the rotation for the first view, and then infer the camera rotation for other views based on relations among cameras.
The rotation among cameras are sampled from priors of multi-view benchmark datasets (\ie, Human3.6M \cite{ionescu2013human3} training set and TotalCapture \cite{Trumble:BMVC:2017} training set).
For camera intrinsic parameters, we fix the focal length and use the center of the rendered images for camera center offset.

\textbf{Data generation.}  
Sampled pose and shape parameters are forwarded into the SMPL model to obtain the vertices of a human mesh. Then we infer $N_\text{J}=17$ COCO 3D joints from the vertices. 
We obtain intermediate representation (\ie, 2D joints and silhouettes) from human mesh and 3D joints via projection and rendering based on Pytorch3D \cite{lassner2020pulsar}. 
Given the sampled camera parameters described before, we project the 3D joints to get 2D joint representation.
Besides, we randomly perturb the vertices $\bm{v}$ to generate a diverse range of human shapes. From perturbed vertices and sampled camera parameters, we render 2D silhouettes using Pytorch3D \cite{ravi2020pytorch3d}. 
We detect the foreground body area based on the union of silhouettes and 2D joints. We crop a bounding box with a scale of $1-1.2$ around the foreground surrounded box, which is unified for consistency between training and testing.   We do zero-padding around the foreground area so that the bounding box is larger than the foreground with a randomly sampled scale from 1 to $1.2$. 
Based on this bounding box crop index, we crop both silhouettes and joints heatmaps and then resize them to the target size ($H=256$, $W=256$) for regression.  Note we also utilize this cropping information for volumetric triangulation during training which is omitted in the equations for simplicity.

\textbf{Data augmentation.}
To simulate noise and discrepancy on 2D joints and silhouettes prediction, we do a series of probabilistic augmentations on each of them independently. Note that the augmentations among different views are sampled independently as well. 
Similar to PartDrop in \cite{zhang2020learning},  we randomly occlude one of the six body parts (head, torso, left/right arm,  left/right leg) on silhouettes with a body-part occlusion probability, and randomly occlude the silhouettes with a box with another predefined box-occlusion probability.
For 2D joints, we randomly perturb the 2D joints position with a deviation and randomly set key joints (\ie, left and right elbow, wrist, knee, ankle) as invisible with a probability. 

% following the hyperparameters in \cite{sengupta2020synthetic} for SMPL pose, shape and camera translation sampling. We use $N_\text{J}=17$ COCO joints  to extract 3D joints from the SMPL model and then project to 2D joint representation. The vertices $\bm{v}$ are randomly perturbed within $[-10\text{mm}, 10\text{mm}]$ for augmentation. 
% From perturbed vertices and sampled camera parameters, we render 2D mask $\bm{M}$ based on Pytorch3D \cite{lassner2020pulsar}. We detect the foreground body area and crop around the foreground area with a scale of $1-1.2$ around the bounding box, which is unified for consistency between training and testing. We crop both human mask $\bm{M}$ and joints heatmaps $\bm{J}$ and then resize to the target size with $H=256$, $W=256$.

\begin{table*}[t]
\centering
% \resizebox{\columnwidth}{!}
\scriptsize
{
\begin{tabular}{ccc}
\toprule
&Hyper-parameter &Value \\ \midrule
\multirow{8}{*}{Sampling} &pose &MoCap priors \\
&{shape $\mu$ } &[0.2056, 0.3356, -0.3507, 0.3561, 0.4175, 0.0309, 0.3048, 0.2361, 0.2091, 0.3121] \\
& shape $\sigma^2$  &$[1.25] \times 10$\\
&camera rotation &MoCap priors, camera priors\\
&camera translation $\mu$ &[0,0,42] m \\
&camera translation $\sigma^2$ &[0.05, 0.05, 5] m\\
&focal length &[5000, 5000] pixel\\ 
\midrule
\multirow{5}{*}{Rendering}
&vertex perturbation  $\mu$  & [0, 0, 0] m \\
&vertex perturbation $\sigma^2$ & [0.01, 0.01, 0] m \\
&rendered image size &(1024, 1024) \\
&bbox scale range &(1.0, 1.2) \\
&regression target size &(256, 256) \\
\midrule
\multirow{7}{*}{Augmentation}
% &bbox center perturbation mean &[0, 0] pixel \\
% &bbox center perturbation variances &[5, 5] pixel \\
&body part occlusion prob. &$[0.1] \times 6$ \\
& box occlusion  prob. &0.1 \\
&  box occlusion dimension &[48, 48]  pixel \\
% & 2D joints L/R swap prob. &0.1 \\
& 2D joints perturbation $\mu$  & $[0] \times 17$\\
& 2D joints perturbation $\sigma^2$ & $[8] \times 17 $\\
& remove 2D joints indices & $[7,8,9,10,13,14,15,16] $\\
& remove 2D joints prob. & $[0.05] \times 8 $\\
\bottomrule
\end{tabular}
}
\caption{Details of hyper-parameters for
training data synthesis and augmentation. }
\label{tab:aug}
% \vspace{-3.3em}
\end{table*}

\textbf{Network and hyper-parameter design.}
We use ResNet-18 \cite{He_2016_CVPR} as encoder where the input is with size $H,W=256$ and the output $\bm{F}$ with size $C=32$, $H_0,W_0=8$. The regression network $\mathcal{R}$ consists of two fully connected layers with 512 neurons each, followed by an output layer with 154 neurons ($\bm{\Theta}=\{ \hat{\bm{\theta}}_\text{j}, \hat{\bm{\beta}}\} \in \mathbb{R}^{23 \times 6 + 10}$). The input of $\mathcal{R}$ are with dimension $ G\times G \times G \times C$ where $G=16$. The length of volumetric cube $L$ is designed as 3 meters. In the loss function, $\omega_\text{j3D}$ and $\omega_\text{v}$ is 1, and all the other weights  are 0.1.
Taking the input vector with dimension $2\times(C_L \times H_0 \times W_0 +3\times H_0\times W_0+2\times N_\text{J})=1540$, the regression network for $\mathcal{R}_\text{fuse}$ consists of two fully connected layers with 1540 neurons each, followed by an output layer with 157 neurons.

\begin{table*}[]
\centering
% \resizebox{\columnwidth}{!}
\scriptsize
{
\begin{tabular}{ccc|ccc|cccc}
\hline
\multicolumn{3}{c|}{Volumetric Mask} &\multicolumn{3}{c|}{Volumetric Fusion}  &\multirow{2}{*}{MPJPE$\downarrow$} &\multirow{2}{*}{PMPJPE$\downarrow$} &\multirow{2}{*}{PVE$\downarrow$} &\multirow{2}{*}{PPVE$\downarrow$}\\
$\bm{\Theta}_1$ &$\bm{\Theta}_2$ &$\bm{\Theta}_3$ &$\bm{\Theta}_1$ &$\bm{\Theta}_2$ &$\bm{\Theta}_3$\\\hline
$\bm{V}_\text{O}^\text{I}$ &- &- &avg &avg &avg &60.0 &48.2 &101.1 &57.7\\
$\bm{V}_\text{O}^\text{I}$ &$\bm{V}_\text{O}^\text{I}$ & $\bm{V}_\text{O}^\text{I}$ &avg &avg &avg  &63.3 &50.6 &107.2 &59.1 \\
$\bm{V}_\text{O}^\text{I}$ &$\bm{V}_\text{O}^\text{U}$  &$\bm{V}_\text{O}^\text{U}$  &avg &avg &avg &57.1 & 45.5 &97.2 &54.0 \\
$\bm{V}_\text{O}^\text{I}$ &$\bm{V}_\text{O}^\text{U}$ &-  &avg &avg &avg &56.1 & 43.1 &92.9 &53.0 \\
$\bm{V}_\text{O}^\text{I}$ &$\bm{V}_\text{O}^\text{U}$ &-  &avg &$\bar{\bm{V}}_{\phi}^n (\bm{\Theta}_1)$ &$\bar{\bm{V}}_{\phi}^n (\bm{\Theta}_2)$ &56.0 & 44.7 &93.1 &54.0 \\
$\bm{V}_\text{O}^\text{I}$ &$\bm{V}_\text{O}^\text{U}$ &-  &avg &avg  &$\bar{\bm{V}}_{\phi}^n (\bm{\Theta}_2)$ &53.8 &42.4 &89.7 &52.3\\ 
\hline
\end{tabular}}
\caption{Ablation study (all in mm) on the Human3.6M dataset (all taking full multi-view images as input). $\bm{V}_\text{O}^\text{U}$ and $\bm{V}_\text{O}^\text{I}$ indicate occupancy union sampling and occupancy intersection sampling respectively (in Equation 9).  $\bar{\bm{V}}_{\phi}^n$ indicates normalized volumetric consistency (in Equation 12). }
\label{tab:h36mab2}
% \vspace{-8mm}
\end{table*}

\section{Ablation Study}
\label{sec:ablations}
\textbf{Ablations on progressive regression.}
In Table \ref{tab:h36mab2} we experiment with different feature sampling and fusion weights on volumetric features $\bm{V}_\text{F}^n$ to regress the mesh $\Theta_1$ , $\Theta_2$ , $\Theta_3$. The results demonstrate the efficiency of our proposed progressive regression from consensus (occupancy union sampling $\bm{V}_\text{O}^\text{U}$), diversity (occupancy intersection sampling $\bm{V}_\text{O}^\text{I}$), and then consistency based balance weighting $\bar{\bm{V}}_{\phi}^n $ for multi-view aggregation.

\textbf{Ablations on canonical space calibration.}
We additionally study the geometric information used in 3D space and reprojection loss as well as  different number of regression iterations. Table \ref{tab:h36mab} shows the results when we employ 1/2/3 iteration of regression and when we use learnable/not learnable parameters as the translation from human center to camera center. We can see from the table that typical iterative regression demonstrates better result than single-iteration regression. Besides, employing estimated human-to-camera translation (world-to-human translation from joints projection) for triangulation in 3D space while employing learnable translation for final reprojection loss achieves faster convergence and better performance.

\begin{table}
\centering
\resizebox{\columnwidth}{!}
% \scriptsize
{
\begin{tabular}{cc|c|cccc}
\toprule
\multicolumn{2}{c|}{3D Triangulation} &{Reprojection} 
&\multirow{2}{*}{MPJPE$\downarrow$} &\multirow{2}{*}{PMPJPE$\downarrow$} &\multirow{2}{*}{PVE$\downarrow$} &\multirow{2}{*}{PPVE$\downarrow$}\\
translation &rotation & loss \\ \midrule
% $\hat\pi^n$ & $\hat{\bm{R}}_{\text{h}\rightarrow \text{c}^n}$  &$\hat\pi^n$ & 1 & 77.5 &59.8 & 123.9 &77.3 
% \\
% $\hat\pi^n$ & $\hat{\bm{R}}_{\text{h}\rightarrow \text{c}^n}$  &$\hat\pi^n$ & 2 &69.2 & 55.1 & 114.9 &70.1 \\
$\hat\pi^n$ & $\hat{\bm{R}}_{\text{h}\rightarrow \text{c}^n}$  &$\hat\pi^n$  &64.1 & 49.0 & 99.7 & 58.2
\\
$\hat{\bm{T}}_{\text{h}\rightarrow \text{c}^n}$  & $\hat{\bm{R}}_{\text{h}\rightarrow \text{c}^n}$  &$\hat{\bm{T}}_{\text{h}\rightarrow \text{c}^n}$, $\bm{K}$ &65.9 &50.1 &108.1 & 63.0
\\
$\hat{\bm{T}}_{\text{h}\rightarrow \text{c}^n}$  & $\hat{\bm{R}}_{\text{h}\rightarrow \text{c}^n}$  &$\hat\pi^n$ &58.1 &45.2 &94.9 &55.8\\ 
\bottomrule
\end{tabular}}
\caption{Ablation study (all in mm) on the Human3.6M dataset. We only employ naive regression to study the efficiency of different calibration design among world space, human space and camera space. $\hat{\bm{T}}_{\text{h}\rightarrow \text{c}^n}$ indicates the estimated translation between human center and world origin based on joints heatmap unprojected in 3D world space. $\hat\pi^n$ indicates the learnable view-specific projection parameters. Since the intrinsic parameters $\bm{K}$ are known, the main difference between $\hat{\bm{T}}_{\text{h}\rightarrow \text{c}^n}$ and $\hat\pi^n$ is that $\hat{\bm{T}}_{\text{h}\rightarrow \text{c}^n}$ is inferred from 2D joints and $\hat\pi^n$ is learned from network.}
\label{tab:h36mab}
% \vspace{-.7em}
\end{table}

\begin{table*}
\centering
\scriptsize
{
\begin{tabular}{ccccccccccc}
\toprule
&Silhouette Perturbation Prob. &0.1 &0.2 &0.3 &0.4 &0.5&0.6&0.7 &0.8 &0.9\\ \hline
\multirow{2}{*}{PPVE$\downarrow$  } 
& wo/ progressive regression  &56.4 &57.0 &57.9 &59.0 &59.6 &60.2 &61.8 & 63.1 &63.9\\
& w/ progressive regression  &52.5 &52.8 &53.3 &53.9 &54.8 &56.0 &57.4 &59.0 &60.3 \\\hline %&117.4 &117.5 &117.8
\multirow{2}{*}{PMPJPE$\downarrow$ } 
& wo/ progressive regression  &45.8 &46.5 &47.7 &49.6 &50.3 &51.8 &53.6 &55.0 &56.9\\
& w/ progressive regression  &42.9 &43.1 &43.7 &44.0 &44.3 &45.1 &45.8 &46.9 &47.8\\
\midrule \midrule %&56.3 &56.4 &56.6
&2D Joints Perturbation Prob.  &0.1 &0.2 &0.3 &0.4 &0.5&0.6&0.7 &0.8 &0.9\\ \hline
\multirow{2}{*}{PPVE$\downarrow$} 
& wo/ progressive regression &56.5 &57.9 &59.2 &61.0 &63.3 &65.0 &67.1 &69.5 &71.2\\
& w/ progressive regression &52.9 &54.0 &55.3 &56.7 &58.4 &60.2 &61.6 &63.1 &65.0\\\hline
\multirow{2}{*}{PMPJPE$\downarrow$}
& wo/ progressive regression &46.3 &49.2 &53.0 &55.7 &59.1 &62.8 &66.0 &69.3 &72.4 \\
& w/ progressive regression &43.2 &46.9 &50.1 &52.9 &56.4 &59.7 &62.8 &65.1 &68.2\\
\bottomrule
\end{tabular}
}
\caption{Comparisons of PVE and PMPJPE (both in mm) when adding random perturbation on silhouette/2D joints representations of Human3.6M test images. We study the performances when using two representations with and without the proposed progressive regression.}
\label{tab:testaug}
\end{table*}

\begin{table}[t]
\centering
% \resizebox{\columnwidth}{!}
\scriptsize
{
\begin{tabular}{c|c|cc}
\toprule
Keypoint-RCNN &DensePose-RCNN &PPVE$\downarrow$ &PMPJPE$\downarrow$ \\\midrule
\multirow{4}{*}{ResNet50\_FPN\_3x} & ResNet50\_FPN &54.0 &42.9\\
& ResNet101\_FPN &54.4 &42.3\\
& ResNet50\_FPN\_DL&53.5 &42.1\\
& ResNet101\_FPN\_DL &53.8 &42.4\\\midrule
ResNet50\_FPN\_1x &\multirow{3}{*}{ResNet101\_FPN\_DL} &53.1 &43.2\\
ResNet50\_FPN\_3x & &53.8 &42.4\\
ResNet101\_FPN\_3x & &53.0 &42.1\\
\bottomrule
\end{tabular}
}
\caption{Comparisons of PPVE and PMPJPE (both in mm) on the Human3.6M dataset when using different model structures for Keypoint-RCNN and DensePose-RCNN to infer 2D joints and silhouettes on Human3.6M test images. ``FPN'' indicates Feature Pyramid Networks.
%``1x'' indicates  training with 12 COCO epochs,  ``3x'' indicates 3x training schedule, and  ``DL'' indicates DeepLabV3 head. 
Note that in this table we report the results testing with progressive regression scheme. }
\label{tab:abdetector}
% \vspace{-1.2cm}
\end{table}

\textbf{Ablations on off-the-shelf detectors.}
We use off-the-shelf detectors to infer 2D joints and silhouettes from RGB images for testing. For 2D joints, we use  pretrained models of Keypoint-RCNN \footnote{https://github.com/facebookresearch/detectron2/blob/main/MODEL\_ZOO.md}. 
For silhouettes, we use pretrained models of
DensePose-RCNN \footnote{https://github.com/facebookresearch/detectron2/blob/main/projects/DensePose\\/doc/DENSEPOSE\_IUV.md}. 
In Table \ref{tab:abdetector}, we compare the results when using different backbones for 2D joints and silhouettes inference. It shows that different model designs result in very minor differences on the 2D joints/silhouettes predictions. The paper reports numbers with the 2D joints/silhouettes inferred with ResNet50\_FPN\_3x for Keypoint-RCNN and ResNet101\_FPN\_DL for DensePose-RCNN.

\textbf{Ablations on noised 2D representation.}
Table \ref{tab:testaug} evaluates the robustness of the proposed method when there is increasing intensity of perturbation (occlusion probability) on 2D joint/silhouette on Human3.6M. We can see that our proposed progressive regression outperforms the baseline (without progressive regression) by a large margin, especially for the cases with severe noise.

\begin{table}[htp]
\centering
\resizebox{\columnwidth}{!}
%\scriptsize
{
\begin{tabular}
{c|c|c|c}
\hline
  &Utility &Supervision &MPJPE(mm) \\ \hline
VoxelPose \cite{tu2020voxelpose} &Pose only &Full &17.8\\ \hline
MvP \cite{zhang2021direct} &Pose only &Full &15.8\\ \hline
Ours &Mesh &Self &22.7\\ \hline
\end{tabular}}
\caption{Comparisons on CMU panoptic dataset. }
\label{tab:cmucomp}
\end{table}
\textbf{Ablations on different camera distributions.}
We further evaluate on CMU Panoptic Dataset \cite{joo2015panoptic} to study our robustness to different camera distributions. We can see from Table \ref{tab:cmucomp} that our method achieves comparable results with the SOTA fully supervised methods while 1) ours does not rely on any prior in domain data; 2) owns additional utility to predict human shape.

\begin{figure*}[t]
    \centering
    \includegraphics[width=1.0\textwidth]{figures/suppfig2.pdf}
    % \vspace{-2em}
    \caption{Visualization of mesh recovery results on Human3.6M, SkiPose and TotalCapture test datasets. We only select four views for visualization on SkiPose and TotalCapture. Yellow-textured meshes are results obtained with baseline methods (naive regression); whereas blue-textured meshes are obtained with our proposed progressive regression scheme.}
    % \vspace{-2em}
    \label{figvis1}
\end{figure*}

\begin{figure*}[t]
    \centering
    \includegraphics[width=1.0\textwidth]{figures/suppfig.pdf}
    % \vspace{-2em}
    \caption{Visualization of mesh recovery results on Human3.6M, SkiPose and TotalCapture test datasets. We only select four views for visualization on SkiPose and TotalCapture. Yellow-textured meshes are results obtained with baseline methods (naive regression); whereas blue-textured meshes are obtained with our proposed progressive regression scheme.}
    % \vspace{-2em}
    \label{figvis2}
\end{figure*}

\section{Qualitative Results}
\label{sec:vis}
In Figure \ref{figvis1} and  \ref{figvis2}, we show qualitative results on baseline method (naive regression with three iterations) and our progressive regression method. The comparisons demonstrate the efficacy of our proposed progressive regression method in multi-view human mesh recovery.

\clearpage

\bibliography{egbib}